\documentclass[10pt]{article}

\usepackage[natbib,style=numeric,backend=biber]{biblatex}

\DeclareBibliographyCategory{math}
\addtocategory{math}{abreu2004beija,arroyo1982community,baird1980selection,barrett1987reproductive,bartomeus2008contrasting,beehler1983frugivory,bek2006pollination,bezerra2009pollination,bluthgen2004bottom, clements1923experimental, dicks2002compartmentalization, dupont2003structure, elberling1999structure, olesen2002invasion, ollerton2003pollination, hocking1968insect, petanidou1991pollination, herrera1988pollination, memmott1999structure, inouye1988pollination, kevan1970high, kato1990insect, medan2002plant, mosquin1967observations, motten1982pollination, primack1983insect, ramirez1992pollination, ramirez1989biologia, schemske1978flowering, small1976insect, smith2005diversity, percival1974floral, ingversen2006plant, philipp2006structure, montero2005ecology, kato2000anthophilous, lundgren2005dense, bundgaard2003tidslig, dupont2009ecological, stald2003struktur, vazquez2002interactions, yamazaki2003flowering, kakutani1990insect, kato1996flowering, kato1993flowering, inoue1990insect, kaiser2010robustness, kaiser2014determinants, robertson1929flowers, vizentin2016influences, del1990hummingbirds, canela2006interaccoes, las2012community, gutierrez2004dinamica, kohler2011redes, lara2006temporal, lasprilla2004interacciones, sabatino2010direct, oppenheimer2021plant, walther2001hummingbird, snow1972feeding, graham2018towards, cotton1998coevolution, buzato2000hummingbird, gonzalez2016species, partida2018pollination, dalsgaard2021influence, rodrigues2014flowers, maruyama2015nectar, carlo2003avian, crome1975ecology, frost1980fruit, snow1971feeding, snow2010birds, galetti2013fruit, hamann1999interactions, jordano1985ciclo, kantak1979observations, lambert1989fig, tutin1997primate, mack1996notes, wheelwright1984tropical, jordano2003invariant, silva2002patterns, guitian1983relaciones, sorensen1981interactions, jordano1993geographical, heleno2013integration, poulin1999interspecific, schleuning2011specialization, davidson1989competition, davidson1991symbiosis, fonseca1996asymmetries, emer2013effects, hadfield2014tale}

\usepackage[toc,page]{appendix}

\AtBeginDocument{%
  \providecommand\BibTeX{{%
    \normalfont B\kern-0.5em{\scshape i\kern-0.25em b}\kern-0.8em\TeX}}}

\usepackage{fullpage}
\usepackage{setspace}
\usepackage{parskip}
\usepackage{titlesec}
\usepackage[section]{placeins}
\usepackage{xcolor}
\usepackage{breakcites}
\usepackage{lineno}
\usepackage{hyphenat}

\PassOptionsToPackage{hyphens}{url}
\usepackage[colorlinks = true,
            linkcolor = blue,
            urlcolor  = blue,
            citecolor = blue,
            anchorcolor = blue]{hyperref}
\usepackage{etoolbox}

\renewenvironment{abstract}
  {{\bfseries\noindent{\abstractname}\par\nobreak}\footnotesize}
  {\bigskip}

\titlespacing{\section}{0pt}{*3}{*1}
\titlespacing{\subsection}{0pt}{*2}{*0.5}
\titlespacing{\subsubsection}{0pt}{*1.5}{0pt}

\usepackage{authblk}

\usepackage{graphicx}
\usepackage[space]{grffile}
\usepackage{latexsym}
\usepackage{textcomp}
\usepackage{longtable}
\usepackage{tabulary}
\usepackage{booktabs,array,multirow}
\usepackage{amsfonts,amsmath,amssymb}
\providecommand\citet{\cite}
\providecommand\citep{\cite}

\newif\iflatexml\latexmlfalse

\AtBeginDocument{\DeclareGraphicsExtensions{.pdf,.PDF,.eps,.EPS,.png,.PNG,.tif,.TIF,.jpg,.JPG,.jpeg,.JPEG}}

\usepackage[utf8]{inputenc}
\usepackage[english]{babel}

\usepackage{float}

\begin{document}

\title{The Robustness of Structural Features in Species Interaction Networks}

\author{Sanaz Hasanzadeh Fard$^1$}
\author{Emily Dolson$^{1,2}$}
\affil{1. Department of Computer Science and Engineering, Michigan State University}
\affil{2. Program in Ecology, Evolution and Behavior, Michigan State University}

\vspace{-1em}

\date{}
  

\begingroup
\let\center\flushleft
\let\endcenter\endflushleft
\maketitle
\endgroup

\selectlanguage{english}
\begin{abstract}

  Species interaction networks are a powerful tool for describing ecological communities; they typically contain nodes representing species, and edges representing interactions between those species. For the purposes of drawing abstract inferences about groups of similar networks, ecologists often use graph topology metrics to summarize structural features. However, gathering the data that underlies these networks is challenging, which can lead to some interactions being missed. Thus, it is important to understand how much different structural metrics are affected by missing data. To address this question, we analyzed a database of 148 real-world bipartite networks representing four different types of species interactions (pollination, host-parasite, plant-ant, and seed-dispersal). For each network, we measured six different topological properties: number of connected components, variance in node betweenness, variance in node PageRank, largest Eigenvalue, the number of non-zero Eigenvalues, and community detection as determined by four different algorithms. We then tested how these properties change as additional edges -- representing data that may have been missed -- are added to the networks. We found substantial variation in how robust different properties were to the missing data. For example, the Clauset-Newman-Moore and Louvain community detection algorithms showed much more gradual change as edges were added than the label propagation and Girvan-Newman algorithms did, suggesting that the former are more robust. Robustness also varied for some metrics based on interaction type. These results provide a foundation for selecting network properties to use when analyzing messy ecological network data.


\end{abstract}%

\sloppy

\section{Introduction}

{\label{874460}}

Graphs are a powerful tool for representing species interactions in community ecology. Species are represented as nodes, while their interactions are represented as edges between nodes. This abstraction is powerful, because it allows graph theory to be harnessed for the purposes of understanding the structure of ecological communities \citep{bascompte2010structure, poisot_describe_2016, delmas_analysing_2019}.
In particular, it is often helpful to calculate graph topology metrics as a way of quantifying the structure of the graph.
These quantifications facilitate answering questions such as 1) what structural patterns a given type of ecological process tends to produce in a species interaction network?, and 2) what ecological processes are predicted by different structural patterns?.

A major obstacle to these approaches, however, is the measurement error inherent in observing real-world species interaction networks \citep{catchen_missing_2023}. It is effectively impossible to exhaustively sample most real-world species interaction networks; some edges are bound to be missed. Too much missing data can render any network topology metric inaccurate. Moreover, systematic bias can be introduced via choice of sampling strategy and techniques for network construction. Likely as a consequence, there is tremendous heterogeneity in the topology of ecological networks \citep{brimacombe_shortcomings_2023}. 

Confronting this problem requires a diversity of approaches. Sampling sufficiency analysis can be conducted to determine how much data is required to get an acceptably high confidence estimate of a given graph topology metric \citep{casas_assessing_2018}. Machine-learning based link-prediction techniques \cite{kumar2020link, fard2023temporal} can be used to infer missing edges in graphs \citep{strydom_roadmap_2021}. Sensitivity analysis can be conducted to infer the severity of the consequences of error in topological metric inference \citep{babtie_topological_2014}. Lastly, new graph metics can be developed that are more robust to common sources of noise in species interaction networks (\textit{e.g.} network size) \citep{botella_appraisal_2022}.

This last option suggests another possible approach: identify existing graph metrics that are less impacted by missing data than others. Prior research in computer science has shown that some graph topology metrics are more robust than others in this respect \citep{zakrzewska_measuring_2014}. Similarly, the aforementioned work on sampling sufficiency revealed that some topological properties require more data to accurately measure than others \citep{casas_assessing_2018}. Thus, we propose that there may be substantial variation in robustness to missing data among graph topology metrics commonly used in ecology, and that it may be possible to exploit this variation to draw more robust conclusions from noisy datasets. Here, we set out to 1) quantify this robustness across a variety of graph topology metrics and 2) propose a technique that others can use to analyze the robustness of their own interaction network topology analysis.

Importantly, how robust a given feature is to missing data will depend on the structure of a network. As an extreme example, imagine that we observe a linear chain network. If we were missing some edges from our sample, we might measure a wildly different distribution of node centrality metrics than we would have on the ground truth network. If, instead, we observe a network with fairly evenly distributed edges, it is less likely that any missing edges will dramatically change the value of centrality metrics. Thus, analysis on the robustness of different graph metrics must be carried out in the context of the network topologies in which we are most interested. 

Here, we perform this analysis for bipartite species interaction networks. The rest of this paper is organized as follows: section $2$ discusses different network properties we evaluated: centrality measures, eigenvalue-related concepts, and community detection algorithms. Section $3$ describes our methods for analyzing the robustness of these measures. Section $4$ brings the results. Section $5$ elaborates on the findings and the last section concludes the paper by future work ideas.

\section{Network Features}
A wide variety of topological features/properties/metrics are used to quantify species interaction network structure. We ultimately chose to analyze the following set: \textit{node centrality} (as measured by betweenness and PageRank), \textit{community detection}  (as measured by the Clauset-Newman-Moore, Louvain, Girvan-Newman, and Label Propagation algorithms), \textit{number of non-zero eigenvalues, number of components, and the largest eigenvalue}. 
Note that we also considered a number of other metrics and chose not to focus on them.
Most importantly, \textit{Connectance and Linkage Density} are both ecologically meaningful properties. However, because they are directly proportional to the number of edges, their robustness to missing edges is entirely predictable from simple math. 
Another interesting metric, \textit{Clustering Coefficient}, is only useful on non-bipartite networks. It would be worthwhile to include in follow-up research on food web properties.

\subsection{Centrality Measures}
Centrality indices rank entities according to their position and importance in the network \citep{brandes2001faster, landherr2010critical, rodrigues2019network, cagua2019keystoneness, gonzalez2010centrality}. Importance can be defined for nodes, links, groups of nodes, and sub-graphs of the network. Here, we focus on node centrality measures. Specifically, we analyze betweenness centrality and PageRank centrality, as these are particularly common approaches. We define these metrics below. It should be mentioned that in this paper, we discuss properties in the case of a simple graph unless otherwise stated.  

\subsubsection{Betweenness Centrality}
In graph theory, the shortest path is defined between a pair of nodes and consists of the smallest number of links that it takes to go from one node to another. The shortest path has applications in ecology, such as predicting how much loss of one species from a community will affect another. Betweenness centrality is a centrality measure defined on the basis of the shortest path. Consider nodes $u$ and $v$ in graph $G$. The shortest path between $u$ and $v$ is the smallest number of links we can trace to go from $v$ to $u$ and vice versa. Betweenness centrality of node $v$ is the number of all pairs of nodes' shortest paths that go through node $v$.  
Equation \ref{first_eqn} shows the betweenness formula.

\begin{equation} \label{first_eqn}
c_B (v) = \sum_{s,t \in V} \frac{\sigma (s,t | v)}{\sigma(s,t)}
\end{equation}

where in Equation \ref{first_eqn}, $V$ is the set of nodes, $\sigma(s,t)$ is the number of shortest (s,t)-paths, and $\sigma(s,t|v)$ is the number of those paths passing through node $v$ where $v \not\in \{s,t\}$. Two special cases occur in this formula: first when $s = t$ so $\sigma(s,t) = 1$, and second when $v \in \{s,t\}$ so $\sigma(s,t|v) = 0$, \citep{brandes2008variants}. 

\subsubsection{PageRank Centrality}
Defined simply, the PageRank algorithm \citep{berkhin2005survey} determines the importance of a node $w$ by counting the number of links to that node. A key underlying assumption is that the more important nodes have more connections. We will provide a quick overview of this algorithm, but note that its details depend heavily on probability theory. We recommend that unfamiliar readers see \citep{durrett2019probability, renyi2007probability} for an overview.

Given a node $w$ of degree $n$ (\textit{i.e.} $n$ neighboring nodes), the PageRank algorithm seeks to find the probability of visiting each of these $n$ neighbors, given that we are currently at node $w$. More formally, given a node, the output of the PageRank algorithm is a probability distribution representing the likelihood of visiting one of its neighbors randomly. In the first step of the PageRank algorithm, all nodes in the graph get the same probability summing up to $1$. 

After visiting a node as the first node (initialization step of the algorithm), in the second step, we can calculate the PageRank of any node $x$, by Equation \ref{second_eqn}.

\begin{equation} \label{second_eqn}
    PR(x) = \sum_{y \in B_x} = \frac{PR(y)}{L(y)}
\end{equation}

In Equation \ref{second_eqn}, $B_{x}$ is the set of all connected nodes to node $x$, and $L(y)$ is the number of links having node $y$ as one of their endpoints.

In the PageRank algorithm, we can set a limit for iterations to control the approximation level and time efficiency.

Here, we use an example to explain how the PageRank algorithm works. Figure $\ref{fig_first}$ represents a graph with $4$ nodes. Figure $\ref{fig_first}$ (left) represents the initialization step where all nodes, $\{w, x, y, z\}$, get the same PageRank, summing up to $1$. In this example, there are $4$ nodes, so the initial PageRank value for each node is $\frac{1}{4}=0.25$. 

\begin{figure}\label{fig_first}
    \centering
    \includegraphics[width=0.8\linewidth]{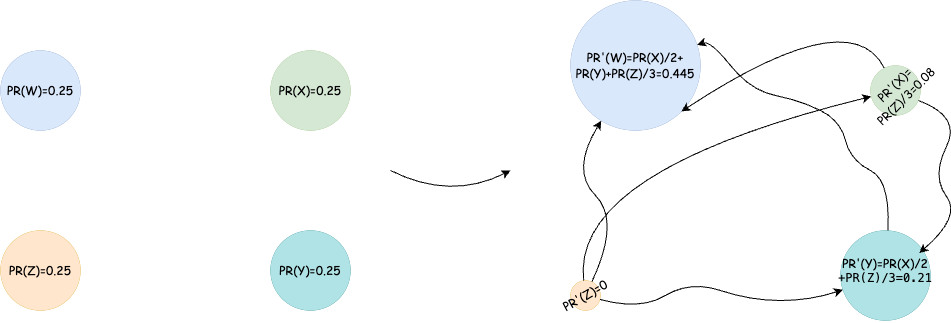}
    \caption{PageRank. Initialization step (left) and final PageRank value after nodes' interactions (right). The size of nodes has a direct relationship with their PageRank value.}
    \label{fig_first}
\end{figure}

Figure $\ref{fig_first}$ (right) represents that node $x$ has links to $w$ and $y$; $y$ has a link to $w$; and $z$ has links to $w, x$, and $y$. The initial PageRank value for all nodes is $0.25$. In the next iteration, node $x$'s PageRank gets divided between $w$ and $y$ equally (each one gets $0.125$). Node $y$'s PageRank gets transferred to $w$ ($0.25$). Node $z$'s PageRank gets divided between $w, x$, and $y$.
As mentioned in Equation \ref{second_eqn}, the amount of PageRank value that some node $a$ transfers to some node $b$, where $a$ has a link to $b$, is equal to $a$'s PageRank divided by the number of nodes that $a$ has links to them. The final PageRank values are given in Figure $\ref{fig_first}$ (right). 




\subsection{Community Detection}
Real-world networks, including biological networks \citep{girvan2002community, alm2003biological}, ecological networks \citep{ings2009ecological, bascompte2010structure}, and social networks \citep{brandes2013social, fard2023machine}, often have community structure \citep{newman2004finding}. Communities are groups of nodes that are more densely connected to each other than to the rest of the network \citep{fortunato2010community}. Detecting these communities \citep{fortunato2010community, fortunato2016community} is one of the most important problems in the study of networks due to their ubiquity, the information they provide about processes that shape a network, and their implications for how future processes will play out on the network. For example, in an ecological context, community detection can reveal groups of species that interact primarily with each other, facilitating abstraction; analysis can then focus on these sub-communities or treat them as higher-level units where appropriate for improved computational tractability. Due to the importance of community detection, a variety of algorithms exist to solve this problem. These algorithms focus on different objectives. For example, some perform better on large networks, others emphasize efficiency over accuracy, and yet others prioritize consistently returning the same result for the same network. 

In this research, we used four different community detection algorithms. The first two, \textit{Clauset-Newman-Moore Community Detection} and \textit{Louvain Community Detection}, are greedy algorithms that detect communities through modularity maximization. The other two algorithms are \textit{Girvan-Newman Community Detection} and \textit{Label Propagation Community Detection}. The Girvan-Newman algorithm is a centrality-based approach and Label Propagation, as implied by its name, finds communities by propagating labels. More detailed summaries of these community detection algorithms can be found in the following sections. All these algorithms solve the problem of assigning nodes to communities.

\subsubsection{Clauset-Newman-Moore Community Detection}

Clauset-Newman-Moore community detection algorithm seeks to maximize the modularity of assigned communities. Modularity is a measure of the quality of a division of a network into communities; community assignments that produce a structure with more links within each community and fewer links between communities have higher modularity. To quantify this property, we will need to perform a few calculations. First, consider $A$ as the adjacency matrix of graph $G$. $A_{vw}$ is an element of matrix $A$ which is $1$ if there is a link between $v$ and $w$, and $0$ otherwise. Also, $c_v$ indicates that node $v$ belongs to community $c$. When considering communities in a graph, there are two types of edges: edges that connect two nodes from the same community, and edges that connect two nodes from two different communities. Equation \ref{third_eqn} reproduced from \citep{clauset2004finding}, counts the number of edges in the first category.

\begin{equation} \label{third_eqn}
    \sum_{vw}{A_{vw}\delta(c_v, c_w)}
\end{equation}

where
\begin{equation} \label{forth_eqn}
    \delta(c_v, c_w) = \begin{cases}
    1 &\text{if $v$ and $w$ are in the same community}\\
    0 &\text{if $v$ and $w$ each belong to a different community}
    \end{cases}
\end{equation}

The number of all links in the graph can be calculated using Equation \ref{fifth_eqn}:
\begin{equation} \label{fifth_eqn}
    m = \frac{1}{2}\sum_{vw}A_{vw}
\end{equation}

Using Equations \ref{forth_eqn} and \ref{fifth_eqn} we compute the fraction of all edges that fall into communities as in Equation \ref{sixth_equation}.

\begin{equation} \label{sixth_equation}
    \frac{\sum_{vw}{A_{vw}\delta(c_v, c_w)}}{\sum_{vw}{A_{vw}}} = \frac{1}{2m}\sum_{vw}{A_{vw}\delta(c_v, c_w)}
\end{equation}

This value can be large when there are many within-community links; but due to the most extreme case, when all nodes and links belong to one community, this value alone cannot be deterministic of the quality of communities. To address this issue, \cite{clauset2004finding} proposed calculating the difference between the value of Equation \ref{sixth_equation} for a particular network and the expected value of Equation \ref{sixth_equation} for a random graph. 

If connections in a graph $G$ are made randomly, the probability of a link existing between $v$ and $w$, of degrees $k_v$ and $k_w$, is equal to $k_vk_w/2m$. Having the above definitions and formulas, \cite{clauset2004finding} defined the modularity measure $Q$ by Equation \ref{fourth_eqn}

\begin{equation} \label{fourth_eqn}
    Q = \frac{1}{2m} \sum_{vw} [A_{vw}-\frac{k_vk_w}{2m}] \delta(c_v,c_w)
\end{equation}

Having the concept and mathematical definition of modularity, we can explain how greedy community detection algorithms mentioned earlier work. The first greedy algorithm we study uses the Clauset-Newman-Moore greedy \citep{clauset2004finding} modularity maximization approach to find the community structure that results in the largest modularity. This algorithm starts by considering each node as one community. In the second step, the algorithm repeatedly merges pairs of communities where this operation results in higher modularity. The algorithm repeats this step until there is no further increase in the modularity. 

\subsubsection{Louvain Community Detection}
We can categorize the Louvain community detection algorithm \citep{blondel2008fast} as a greedy, modularity-based community detection algorithm. It is a heuristic algorithm that tries to find the best partitions of the graph by heuristically optimizing modularity. Similar to the greedy community detection algorithm, the Louvain algorithm starts by assigning each node to be in its own community. Then, the algorithm iterates over the existing communities. For each community, the algorithm tests the effect of joining that community with each of its neighboring communities on modularity. If none of these options increases the overall modularity score, the community is left as it is. Otherwise, it gets joined with the neighboring community which increases the modularity score the most. This process continues until there is no individual move left that can increase the modularity. 

In the second step, a new graph is built. The nodes of this graph are the communities we found in the first step. The process of building a new graph (steps $1$ and $2$) continues until no further increase in modularity is possible. 

\subsubsection{Girvan-Newman Community Detection}
Girvan-Newman \citep{girvan2002community} is another community detection algorithm. This algorithm is fundamentally different from greedy community detection algorithms. Greedy community detection algorithms function based on modularity maximization while the Girvan-Newman algorithm is a centrality-based algorithm. At each step, the edge with the highest betweenness centrality is removed, partitioning the community into two sub-communities. This process is repeated until modularity stops increasing.

\subsubsection{Label Propagation Community Detection}
Label propagation is a semi-supervised machine-learning algorithm for clustering. In the case of community detection, this algorithm assigns labels to nodes. 

Similarly to the formerly mentioned algorithms, label propagation starts by labeling each node as its own community. In the next step, the label of each node is updated to be the label most common among its immediate neighbors. 
The label propagation algorithm has the advantage of being fast. 
One drawback of this algorithm is that it is not guaranteed to always produce the same solution. 

Figure \ref{all_com} represents the output of each community detection algorithm on the same graph.

\begin{figure} \label{all_com}
    \centering
    \includegraphics[width=1\linewidth]{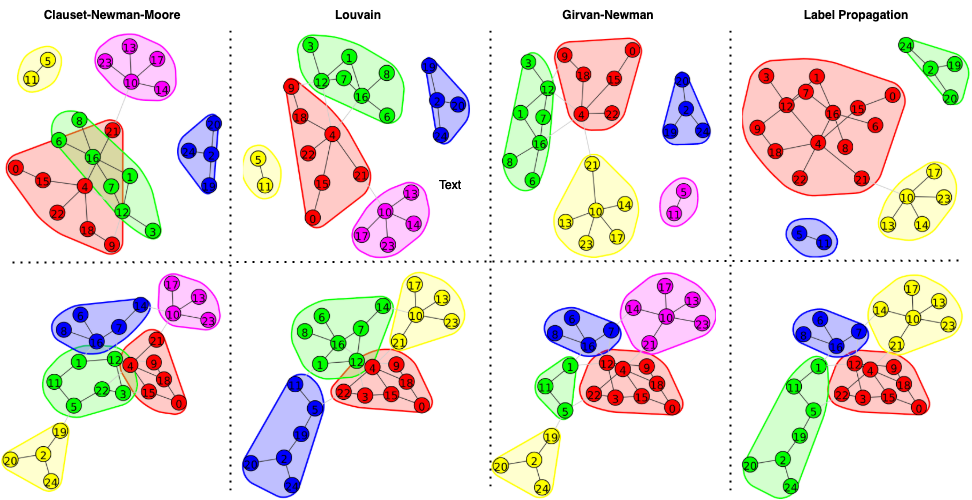}
    \caption{Performance of different community detection algorithms on the same graph; detected communities in the graph (up), detected communities after adding 7 random edges (down).}
    \label{all_com}
\end{figure}

\subsection{Eigenvalues}
Eigenvalues are a powerful tool for analyzing graphs that provide information based on solid mathematical proof. 
In order to explain eigenvalue-based analysis, we must first introduce the concepts of adjacency matrices, degree matrices, and graph \textit{Laplacians}. Adjacency matrices are a common way to represent graphs in computers. The adjacency matrix is a matrix with a row and column for each node in the graph. In the case of a simple graph, the adjacency matrix is symmetric. Consider a graph $G$, where $V = \{a, b, c, d, e\}$ is the set of nodes and $E = \{(a, b), (a, d), (b, c), (b, e), (c, d), (c, e)\}$ is the set of links of the network (see Figure \ref{second_fig} (1)). Matrix $A$ (see Figure \ref{second_fig} (2)) is the adjacency matrix of graph $G$. The entry at row $i$ and column $j$ where $\{i, j\} \in V$ is $1$ (or the edges' weight if the graph is weighted) if there is a link between $i$ and $j$, and otherwise it is $0$.

The degree matrix, $D(G)$ (see Figure \ref{second_fig} (3)), has the same dimension as the adjacency matrix. All the entries of $D(G)$ are $0$ except for the diagonal ones. In the degree matrix the entry $(i, i)$ where $i \in V$ is equal to $i$'s degree.

We represent the Laplacian of the graph $G$ as $L(G)$, shown in Figure \ref{second_fig} (4). The Laplacian of $G$ is the difference between the degree matrix and the adjacency matrix \citep{merris1998laplacian}. This matrix is the one that we will actually calculate its eigenvalues for the purposes of graph analysis. In particular, the Laplacian can be used to study the behavior of the network when links get added or deleted or nodes get merged \citep{merris1998laplacian}.

\begin{figure}\label{second_fig}
    \centering
    \includegraphics[width=0.8\linewidth]{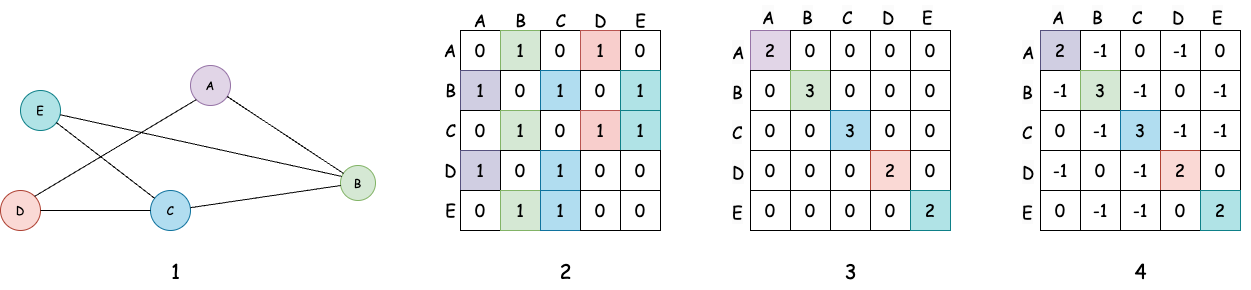}
    \caption{Graph (1), Adjacency Matrix (2), Degree Matrix (3), Laplacian Matrix (4)}
    \label{second_fig}
\end{figure}

In this research, we studied the number of $L(G)$'s non-zero eigenvalues, the largest eigenvalue, and the number of zero eigenvalues, which indicates the number of connected components of the graph.

\section{Experiments}
We seek to quantify the response of network properties to varied levels of missing data. To do so, we first identified a set of ecological interaction networks measured in the real world. We then simulated potential ground truth networks that these measured networks could potentially have been sampled from.

To reduce the effect of randomness on results, we analyzed 148 networks from the Web of Life project \citep{fortuna2014web}.
These networks are gathered from the literature. Table \ref{first_table} represents the statistics of the networks analyzed here (see Appendix \ref{A} for the related resources).

\begin{table}[] \label{first_table}
\centering
\caption{Statistics of the analyzed networks}
\begin{tabular}{@{}lc@{}}
\toprule
\multicolumn{2}{c}{\textbf{DATABASE STATISTICS}}                      \\ \midrule
\textbf{\# Networks}          & 148                            \\
\textbf{\# Species}           & 6 - 1,500                           \\
\textbf{\# Interactions}      & 6 - 15,255                           \\
\textbf{Connectance}          & 0.017 - 0.688                                 \\
\textbf{Type of Interactions} & Pollination, Host-Parasite, Plant-Ant, Seed-Dispersal \\ \bottomrule
\end{tabular}
\label{first_table}
\end{table}

It is possible that different types of interaction network may behave differently from each other in a systematic way. To determine whether this is happening, we selected four types of interaction network from the Web of Life database to analyze: pollination, plant-ant, host-parasite, and seed-dispersal.

We selected these types because they represented substantial diversity in ecological interactions while 1) being bipartite, 2) being large enough for our analysis to be possible, and 3) having enough networks to potentially be able to make useful generalizations.
All graphs were bipartite, meaning that each had $2$ partitions of nodes, where nodes in one partition have no interactions and every interaction occurs between two nodes from different partitions (e.g. pollinators only interact with plants, not with other pollinators). Because most graphs in the dataset were unweighted, we did not consider edge weight in our analysis.

Any time a species interaction network is measured in nature, there is a chance of failing to detect an interaction that is present. Consequently, missing edges may or may not indicate a true of lack of interaction. 
Thus, we can assume that any observed network contains a set of real interactions. However, it also is missing some edges relative to the theoretical ``ground truth'' network, which we define as the true set of all interactions between species in the network.
While we can never know what the ground truth network looked like, we can simulate possible sets of additional edges that it may contain relative to the observed graph.
Graph properties that are more robust to missing interactions will be more similar between these simulated ground truth networks and observed graphs.
To test whether features are robust, then, we can randomly generate graphs that the observed networks could have been sampled from and compare the values of graph properties between them.

We generated simulated ground truth graphs based on the observed graphs in the Web of Life datasets using the following process. Consider a graph \textit{G}, with a set of vertices $V$, and a set of edges \textit{E}, where $|E| = m$. We generate candidate ground truth graphs $G' = \{g'_1, g'_2, ..., g'_{m/2}\}$, with the same set of vertices as $G$ ($V$) and a set of edges equal to $E$ plus $\{1, 2, ..., m/2\}$ randomly generated added edges. The maximum number of missing edges we consider is $m/2$, meaning that in the most extreme case, \textit{g'} has $m+m/2$ edges. For each observed graph and number of added edges, we generate 10 candidate graphs representing possible ground truths. We then conduct our analyses over this collection of graphs. 

\section{Results}

Figure \ref{fourth_fig} (left) shows the results for the number of non-zero eigenvalues. 
Generally, dense graphs have more non-zero eigenvalues than sparse ones with the same number of nodes. Thus, we would expect the number of non-zero eigenvalues to increase as we add edges. Indeed, as shown in Figure \ref{fourth_fig} (left), this value is strictly increasing from left to right. 
Figure \ref{fourth_fig} (right) represents the number of components in terms of the number of zero eigenvalues; we can see in Figure \ref{fourth_fig} (right) by moving toward the right side of the plot and having more links in the graphs, the number of components decreases. 

The central question, then, is what the pattern of increase in the number of non-zero eigenvalues as we add edges looks like. In general, the number of non-zero eigenvalues is a fairly robust measure across all four types of networks that we examined (see Figure \ref{fourth_fig} (left)). In particular, for host-parasite and seed-dispersal networks, adding edges only slightly increases the value of this metric. This metric is also robust for pollination and plant-ant networks up to a point; when we add fewer than a quarter of the possible missing edges, the number of non-zero eigenvalues stays very consistent. Past that point, however, it does change more dramatically, particularly for plant-ant networks. 

Figure \ref{fourth_fig} (right) shows the results for the number of connected components in the graph. Adding edges to a graph can only decrease the number of connected components, and the minimum number of connected components in a graph is one. These constraints are born out in the data.
For all datasets, the number of components converges to one after adding even a few edges.
This discrepancy is particularly large for the plant-ant dataset, as the observed networks in it had a relatively large number of connected components on average.
For other network types, converging to a single connected component represents a smaller change.
Thus, for plant-ant networks, there is a high risk that missing data could dramatically change the value of this metric. 


The number of communities detected by community detection algorithms is, in general, a less robust metric than the number of non-zero eigenvalues. Among community detection algorithms however, the Clauset-Newman-Moore and Louvain algorithms produce a relatively robust measure of the number of communities (see Figure \ref{fifth_fig} (top-right) and (top-left)). As expected, across all community detection algorithms, adding links usually resulted in a smaller number of communities detected. In the Label Propagation algorithm, this drop occurs rapidly, even at a small number of added edges (see Figure \ref{fifth_fig} (bottom-left)). Indeed, with only a few additional edges, the Label Propagation count of communities drops all the way to one. In contrast, the Clauset-Newman-Moore and Louvain community counts drop more gradually as edges are added. For the most part, there is not a dramatic qualitative difference across network types.  

The Girvan-Newman community detection algorithm is somewhat hard to classify in terms of robustness (see Figure \ref{fifth_fig} (bottom-right)). In general, it detects fewer communities in the observed graphs than the other algorithms. Moreover, adding even a small number of edges consistently causes the number of detected communities to drop to two. In some sense, this behavior could be considered robust (for network types other than Plant-Ant), because there is relatively little change as edges are added. However, the discrepancy in the results of the other algorithms is concerning.


Figure \ref{six_fig} shows the results from two centrality measures: betweenness and PageRank variance. Once a relatively small number of edges have been added, these measures show similar behavior and the only difference is for the plant-ant network. Once we've added a quarter or more of the possible missing edges, values of both betweenness and PageRank variance fall to a consistent low level. This behavior makes sense, as adding edges will generally reduce the variance in centrality.


The largest eigenvalue, as shown in Figure \ref{sixth_fig} consistently changes linearly until we have added half of the possible missing edges. From there, for most network types, it levels off and stays fairly constant. The one exception is pollination networks, which keep changing mostly linearly.

\section{Discussion}

The results presented here are a tool for researchers to use in selecting network features to measure. The properties we analyzed exhibited a diversity of patterns of response to added edges. Some of these patterns showed consistent gradual change, indicating that the amount of inaccuracy in the metric would likely be proportional to the amount of edges incorrectly missing from the graph. Other metrics exhibited immediate dramatic change. Using these metrics in cases where there is a risk of much error in network topology should only be done with great caution, as small errors in underlying data could produce large errors in resulting measurements. Finally, same patterns were in between; using them on networks with a small number of missing edges should not produce serious inaccuracies in the final results, but using them with networks missing more edges could. In general, researchers should consider both the amount of error in their data and the robustness profile of the metrics they are interested in when making experimental design choices.

Among the features we studied, based on the results, the number of non-zero eigenvalues (see Figure \ref{fourth_fig} (left)) is the most reliable measure in the presence of missing data. Across network types, this value stays fairly constant as we add edges. It is important as we can infer many facts about the network by transforming it into Laplacian space; knowing the network features remain robust in this space can help with a range of analysis. 

The robustness of community detection algorithms varied substantially. The Clauset-Newman-Moore and Louvain algorithms both showed fairly gentle and continuous change as edges were added, meaning they are fairly robust to missing data. In contrast, the label propagation and Girvan-Newman showed dramatic change in response to the addition of a relatively small number of edges.

This lack of robustness may be due to the size of the networks, as some community detection algorithms are designed for very large networks and do not perform well on smaller networks.

These results illustrate the fact that there can be substantial variation in robustness to missing data across techniques for measuring the same graph property. Researchers should consider the expected amount of error in their data when selecting an algorithm. In particular, these results suggest favoring the Clauset-Newman-Moore and Louvain algorithms in cases where there is a high risk of failing to detect edges.

\section{Conclusion and Future Work}
In this study, we have proposed a new approach to strengthening the analysis of species interaction network structure in light of the potential for missing data. We analyzed a range of network features under varied levels of simulated missing data and found substantial variation in their robustness. Some properties changed dramatically when even a small number of edges were added, while others changed more gradually. This effect varied by network type in some cases, supporting our expectation that metric robustness varies based on graph topology.  
Indeed, while we expect many of our results to generalize across bipartite species interaction networks, we also see our methodology as a key contribution of this work. Given the uncertainty inherent in most observed interaction network data, researchers quantifying network topology could consider performing sensitivity analyses by testing the impact of adding or removing edges.  

There are of course some important limitations to this work. This technique will not account for systematic bias in the data collected, although it could be adjusted to do so by simulating potential biases in the edge addition process. Similarly, the assumption that all edges are equally likely is an oversimplification; fortunately, though, it should mostly bias graph metrics towards looking less robust than they really are. This oversimplification could be corrected with more sophisticated link prediction algorithms. Lastly, there are many additional types of network and graph property that this analysis could be extended to. It would be interesting to see how different these results are in non-bipartite networks, such as food webs.

Overall, this work adds another tool to our toolbox for doing robust analysis of species interaction networks, despite the messiness of the underlying data. Given recent concerns about the utility of analyzing the topology of networks in large databases \citep{brimacombe_shortcomings_2023}, we hope this technique can provide a path forward for conducting large scale network ecology research.



\begin{figure}\label{fourth_fig}
    \centering
    \includegraphics[width=0.8\linewidth]{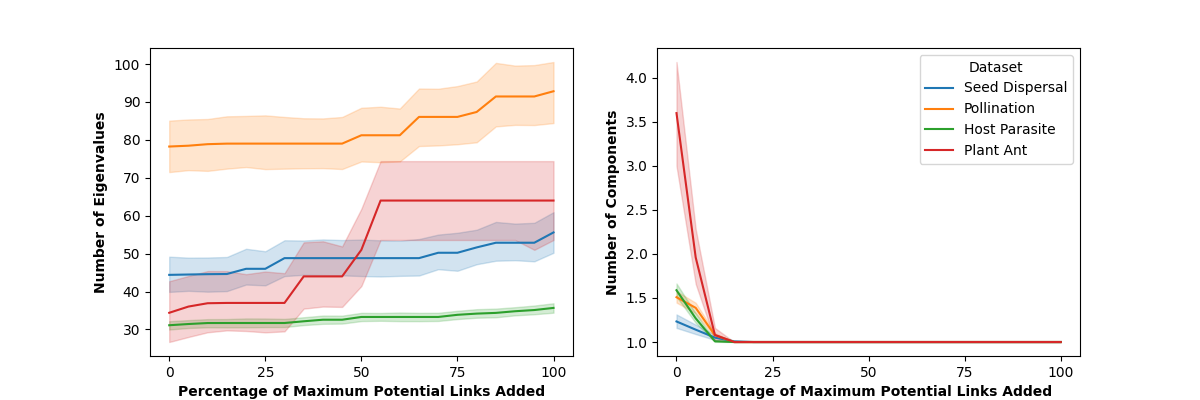}
    \caption{The effect of missing data on (left) the number of non-zero eigenvalues, and (right) the number of connected components detected by the number of zero eigenvalues.}
    \label{fourth_fig}
\end{figure}

\begin{figure}\label{fifth_fig}
    \centering
    \includegraphics[width=0.8\linewidth]{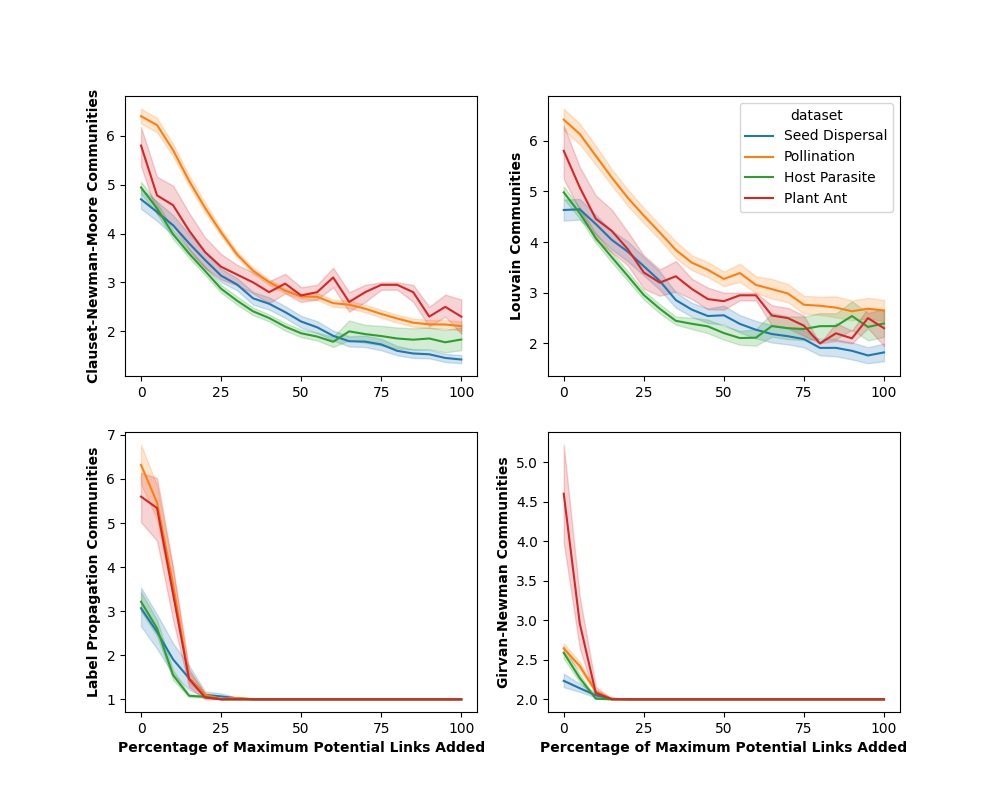}
    \caption{The effect of missing data on the number of communities detected by left to right and from top to bottom (1) Clauset-Newman-Moore algorithm, (2) Louvain community detection algorithm, (3) label propagation algorithm, and (4) Girvan-Newman algorithm.}
    \label{fifth_fig}
\end{figure}

\begin{figure}\label{six_fig}
    \centering
    \includegraphics[width=1\linewidth]{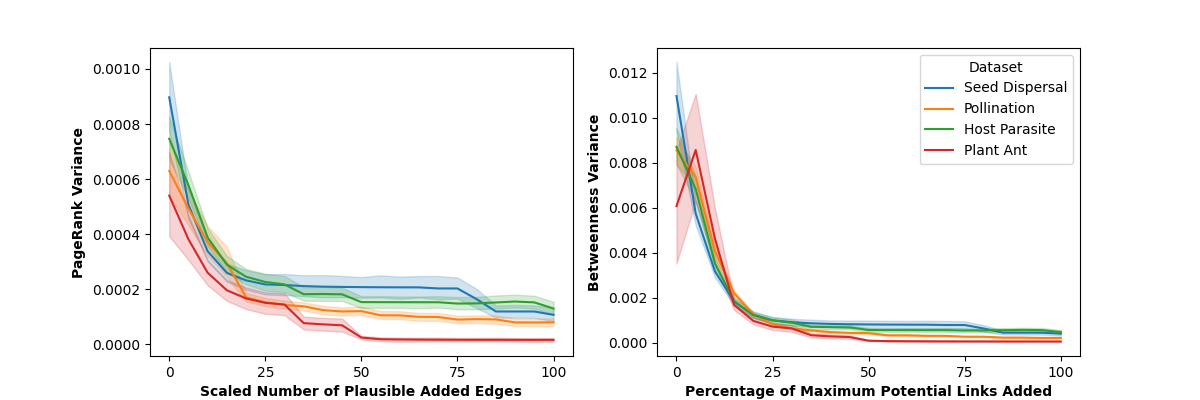}
    \caption{The effect of missing data on (left) PgeRank variance and (right) betweenness variance.}
    \label{six_fig}
\end{figure}

\begin{figure}\label{sixth_fig}
    \centering
    \includegraphics[width=0.5\linewidth]{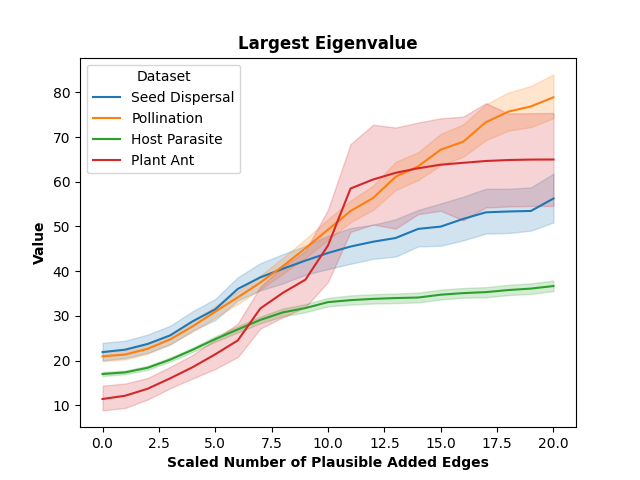}
    \caption{The effect of missing data on the largest eigenvalue}
    \label{sixth_fig}
\end{figure}

\newpage






\printbibliography[notcategory={math}, resetnumbers=true]

\nocite{*}
\newpage

 \begin{appendices}


The networks we used for this study are sourced from 
\cite{abreu2004beija,arroyo1982community,baird1980selection,barrett1987reproductive,bartomeus2008contrasting,beehler1983frugivory,bek2006pollination,bezerra2009pollination,bluthgen2004bottom, clements1923experimental, dicks2002compartmentalization, dupont2003structure, elberling1999structure, olesen2002invasion, ollerton2003pollination, hocking1968insect, petanidou1991pollination, herrera1988pollination, memmott1999structure, inouye1988pollination, kevan1970high, kato1990insect, medan2002plant, mosquin1967observations, motten1982pollination, primack1983insect, ramirez1992pollination, ramirez1989biologia, schemske1978flowering, small1976insect, smith2005diversity, percival1974floral, ingversen2006plant, philipp2006structure, montero2005ecology, kato2000anthophilous, lundgren2005dense, bundgaard2003tidslig, dupont2009ecological, stald2003struktur, vazquez2002interactions, yamazaki2003flowering, kakutani1990insect, kato1996flowering, kato1993flowering, inoue1990insect, kaiser2010robustness, kaiser2014determinants, robertson1929flowers, vizentin2016influences, del1990hummingbirds, canela2006interaccoes, las2012community, gutierrez2004dinamica, kohler2011redes, lara2006temporal, lasprilla2004interacciones, sabatino2010direct, oppenheimer2021plant, walther2001hummingbird, snow1972feeding, graham2018towards, cotton1998coevolution, buzato2000hummingbird, gonzalez2016species, partida2018pollination, dalsgaard2021influence, rodrigues2014flowers, maruyama2015nectar, carlo2003avian, crome1975ecology, frost1980fruit, snow1971feeding, snow2010birds, galetti2013fruit, hamann1999interactions, jordano1985ciclo, kantak1979observations, lambert1989fig, tutin1997primate, mack1996notes, wheelwright1984tropical, jordano2003invariant, silva2002patterns, guitian1983relaciones, sorensen1981interactions, jordano1993geographical, heleno2013integration, poulin1999interspecific, schleuning2011specialization, davidson1989competition, davidson1991symbiosis, fonseca1996asymmetries, emer2013effects, hadfield2014tale}.

\printbibliography[category={math},title={Appendices References}, resetnumbers=true]

\end{appendices}

\end{document}